\documentclass[acmtog,authorversion]{acmart} 

\usepackage{booktabs}
\usepackage{graphicx}
\usepackage{enumitem}
\usepackage{soul}
\usepackage{dsfont}
\usepackage{gensymb}
\usepackage{microtype}
\usepackage{pifont}

\usepackage{amssymb}
\usepackage{amsmath}
\usepackage{multirow}	
\usepackage{color, colortbl}
\usepackage{collect}
\usepackage{xspace}
\usepackage{makecell}
\usepackage[capitalize,nameinlink]{cleveref}
\usepackage{xcolor}

\definecolor{changedcolor}{RGB}{0, 0, 0}   
\newcommand{\changed}[1]{{\color{changedcolor}#1}}

\colorlet{bestcolor}{blue!30}
\colorlet{secondbestcolor}{blue!15}
\colorlet{thirdbestcolor}{gray!15}

\newcommand{\best}[1]{\cellcolor{bestcolor}#1}
\newcommand{\second}[1]{\cellcolor{secondbestcolor}#1}
\newcommand{\third}[1]{\cellcolor{thirdbestcolor}#1}

\begin{document}

\title{Faster 3D Gaussian Splatting Convergence via Structure-Aware Densification}

\author{Linjie Lyu}
\email{llyu@mpi-inf.mpg.de}
\orcid{0009-0007-4763-8457}
\affiliation{%
  \institution{Max-Planck-Institut für Informatik}
  \country{Germany}
}
\author{Ayush Tewari}
\email{at2164@cam.ac.uk}
\orcid{0000-0002-3805-4421}
\affiliation{%
  \institution{Cambridge University}
  \country{United Kingdom}
}

\author{Jianchun Chen}
\email{jchen@mpi-inf.mpg.de}
\orcid{0009-0001-8673-646X}
\affiliation{%
  \institution{Max-Planck-Institut für Informatik}
  \country{Germany}
}

\author{Thomas Leimkühler}
\email{thomas.leimkuehler@mpi-inf.mpg.de}
\orcid{0009-0006-7784-7957}
\affiliation{%
  \institution{Max-Planck-Institut für Informatik}
  \country{Germany}
}

\author{Christian Theobalt}
\email{theobalt@mpi-inf.mpg.de}
\orcid{0000-0001-6104-6625}
\affiliation{%
  \institution{Max-Planck-Institut für Informatik}
  \country{Germany}
}
\affiliation{%
  \institution{Saarbrücken Research Center for Visual Computing, Interaction, and Artificial Intelligence (VIA)}
  \country{Germany}
}

\begin{abstract}
3D Gaussian Splatting has emerged as a powerful scene representation for real-time novel-view synthesis. However, its standard adaptive density control relies on screen-space positional gradients, which do not distinguish between geometric misplacement and frequency aliasing, often leading to either over-blurred high-frequency textures or inefficient over-densification. We present a structure-aware densification framework. Our key insight is that the decision to subdivide a Gaussian should be driven by an explicit comparison between its projected screen-space extent and the local structure of the texture it seeks to represent. We introduce a multi-scale frequency analysis combining structure tensors with Laplacian scale space analysis to estimate the dominant frequency at each pixel, enabling robust supervision across varying texture scales. Based on this analysis, we define $\eta$, a per-Gaussian, per-axis frequency violation metric that indicates when a primitive may be under-resolving local texture details. Unlike methods that perform isotropic splitting (e.g., splitting each Gaussian into two smaller ones with uniform shape), our approach performs anisotropic splitting. For each axis with high $\eta$, we compute a split factor to better resolve the local frequency content. We further introduce a multiview consistency criterion that aggregates $\eta$ observations across multiple views. By performing densification early and faster, we skip the lengthy iterative densification phases required by baseline methods and achieve significantly faster convergence. Experiments on standard benchmarks demonstrate that our method also achieves superior reconstruction quality, particularly in high-frequency regions.\looseness=-1
\end{abstract}

\begin{teaserfigure}
  \includegraphics[width=\textwidth]{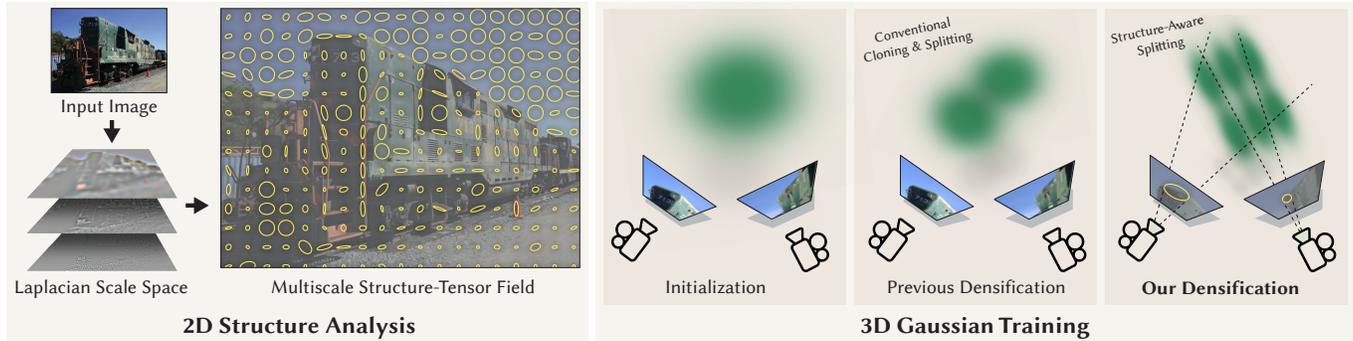}
  \vspace{-6mm}
  \caption{Our approach addresses population control in the 3D Gaussian Splatting representation. Instead of conventional split or clone operations for Gaussians, we analyze multiscale image structure in the input views (left) and leverage this information for structure-aware anisotropic densification (right). Our method leads to significant acceleration of convergence (53\,\text{s} on the Mip-NeRF360 and 41\,\text{s} on the Deep Blending dataset) and achieves the best perceptual quality.}
  \label{fig:teaser}
  \Description{}
\end{teaserfigure}



\maketitle

\section{Introduction}
\label{sec:intro}

3D Gaussian Splatting (3DGS)~\cite{kerbl20233dgs} has emerged as an effective approach for novel-view synthesis, achieving real-time rendering with high-fidelity reconstruction. Unlike implicit neural representations such as NeRF \cite{mildenhall2020nerf}, 3DGS explicitly represents scenes as collections of anisotropic 3D Gaussians, enabling efficient tile-based rasterization. However, despite its rendering efficiency, training a 3DGS model typically requires 30,000 iterations that can take 10--20 minutes even on modern GPUs. 

A common assumption is that the training bottleneck lies in the optimization process itself: the learning rate schedule, gradient computation~\cite {mallick2024taming3dgs}, or convergence properties of the optimizer~\cite {hollein20253dgs,lan20253dgs2}. However, we argue that \emph{one of the main bottlenecks is densification}. The standard adaptive density control in 3DGS relies on a reactive, gradient-based heuristic that splits primitives only after significant rendering error has accumulated. Furthermore, an isotropic split strategy is applied, where each split operation divides one Gaussian into just two children.  Regions with high-frequency details require a dense sampling of Gaussians, which can only be achieved by repeating this process many times, each round requiring hundreds of training iterations to accumulate sufficient gradient statistics before the next split can occur.
As an example, consider a Gaussian that needs to resolve a texture requiring 16 times its current sampling density. Under the traditional scheme, this requires at least four successive split-and-train cycles ($2^4 = 16$), each spanning hundreds of iterations. If we could determine the required resolution analytically and densify accordingly in a single step, we could bypass the lengthy iterative densification entirely.

The gradient-based splitting criterion is fundamentally \textit{uninformed} about the local scene structure. A Gaussian that perfectly covers a textured region but is too large to resolve fine details produces a blurry reconstruction, yet may generate small positional gradients because its \textit{center} is already well-placed. The standard heuristic struggles to distinguish between geometric misalignment (where the Gaussian should move) and inadequate resolution (where it should split). In practice, finding an appropriate gradient threshold to trigger splitting for thin structures while avoiding over-densification elsewhere remains a challenging balancing act. 
Previous methods have explored advanced adaptive densification criteria~\cite {mallick2024taming3dgs,ren2026fastgs} to reduce the Gaussian count and speed up individual iterations; however, they remain fundamentally limited by the traditional isotropic splitting strategy, which fails to efficiently capture thin geometric structures and bottlenecks convergence speed through gradual densification schedules.

In this work, we propose a structure-aware densification framework that directly addresses the densification bottleneck. We introduce a supervision signal based on \textit{structure tensors} \cite{di1986note,forstner1987fast} combined with \textit{Laplacian scale space} analysis~\cite{lindeberg1994scale}, which together enable robust estimation of the dominant frequency and orientation of image features at each pixel. This multi-scale approach identifies the characteristic frequency across multiple octaves, making the method robust to textures of varying scales.

By projecting each Gaussian's 3D axes into screen space and comparing the resulting extents against the local texture wavelength, we derive a per-axis \textit{frequency violation metric}, $\eta$. When $\eta > 1$, the primitive may be too large to faithfully represent the local texture, indicating that subdivision could improve detail reconstruction.

Unlike previous methods that perform stochastic sampling-based splitting (typically generating children via random offsets and shrinking all axes uniformly), our approach utilizes $\eta$ to perform \emph{early analytic densification}. We calculate the number of subdivisions $n$ required along each axis independently, generating $n_x \times n_y \times n_z$ children arranged in a regular grid. This allows our method to skip the gradual and long densification periods required by reactive gradient-based methods, ensuring that the new primitives are immediately capable of resolving the local frequency content. 
This leads to faster convergence and better detail reconstruction.

Furthermore, we introduce a \textit{multiview consistency mechanism} that aggregates $\eta$-observations across multiple training views. Rather than reacting to errors from a single viewpoint, we classify each Gaussian's observations and trigger densification only when a significant fraction of views report consistent frequency violations. This prevents over-densification for non-surface Gaussians. 

In summary, our contributions are summarized as follows:
\begin{enumerate}[leftmargin=*]
    \item A \textbf{Multi-scale Frequency Analysis} utilizing structure tensors and Laplacian scale space to compute the dominant frequency at each pixel, providing robust supervision across texture scales.\looseness=-1
    
    \item A \textbf{Frequency Violation Metric} that quantifies potential under-resolution by comparing projected Gaussian extent to local texture wavelength.
    
    \item An \textbf{Structure-aware Densification} strategy that determines the optimal per-axis subdivision factor  to resolve frequency violations efficiently, enabling the model to skip lengthy densification periods.
   
\end{enumerate}

Experiments on standard benchmarks demonstrate that our method achieves superior reconstruction quality, particularly in high-frequency regions such as fine structures, while also improving training efficiency over baseline methods.
\changed{All code and data are available on our project webpage: \url{https://vcai.mpi-inf.mpg.de/projects/SAD-GS}.}

\section{Related Work}
\label{sec:related}

Here, we review related works on 3DGS, focusing on acceleration techniques, densification strategies, and frequency analysis.

\paragraph{3DGS Acceleration.}
A key bottleneck of 3DGS~\cite{kerbl20233dgs} is the large number of Gaussians required for high-fidelity reconstruction, which increases both training time and rendering cost. Several works focus on optimizing the rasterization pipeline itself. Taming-3DGS \cite{mallick2024taming3dgs} replaces per-pixel backpropagation with per-splat parallel computation, significantly speeding up optimization and serving as a strong baseline for subsequent research. StopThePop \cite{radl2024stopthepop} and FlashGS \cite{Feng_2025_CVPR} use precise tile intersection to reduce Gaussian--tile pairs and accelerate rasterization. 3DGS-LM \cite{hollein20253dgs} replaces the Adam~\cite{kingma2014adam} optimizer with the Levenberg-Marquardt algorithm for faster convergence, and 3DGS$^2$ \cite{lan20253dgs2} achieves near second-order convergence via prioritized per-kernel updates. These methods primarily target computational efficiency rather than addressing the fundamental question of \textit{which} Gaussians to create or remove.

\paragraph{Gaussian Densification and Pruning.}
The quality of 3DGS reconstructions depends heavily on adaptive density control, which clones or splits Gaussians based on view-space positional gradients. Recent works refine this densification strategy to control primitive count. Taming-3DGS \cite{mallick2024taming3dgs} employs a budgeting mechanism using importance scores based on Gaussian-associated properties. DashGaussian \cite{chen2025dashgaussian} introduces a resolution-guided scheduler that progressively reconstructs the scene. Compact-3DGS \cite{lee2024compact} uses learnable masks to reduce redundancy. Similarly, \citet{kheradmand20243d} reformulate 3DGS optimization as a Markov Chain Monte Carlo process to dynamically adjust Gaussian density. Nevertheless, these methods still require a prolonged densification phase when relying on the gradient-based heuristic.

Beyond densification, several methods focus on pruning to reduce Gaussian count. Mini-Splatting \cite{fang2024minisplatting} removes Gaussians through intersection-preserving simplification and sampling. PUP 3DGS \cite{hanson2025pup} and Speedy-Splat \cite{hanson2025speedy} remove redundant Gaussians by computing Hessian approximations across training views. LightGaussian \cite{fan2024lightgaussian} and Compact3D \cite{navaneet2023compact3d} design importance scores to guide pruning. While effective at reducing primitive count, these methods may discard Gaussians needed to capture high-frequency details, degrading rendering quality.  FastGS \cite{ren2026fastgs} introduces a multiview consistent importance score for densification and pruning. We consider this work to be our state-of-the-art baseline. FastGS is limited by a gradual densification schedule that bottlenecks convergence speed; in contrast, we demonstrate that structure-aware early densification enables significantly faster convergence.

\paragraph{Frequency Analysis in Radiance Fields.} Aliasing is a fundamental challenge in rendering, arising when the sampling rate is insufficient to capture high-frequency scene content. In NeRF-based methods, Mip-NeRF \cite{barron2021mipnerf} introduced cone tracing and integrated positional encoding to pre-filter the representation.

For 3D Gaussian Splatting, recent works utilize frequency analysis to guide the structural distribution of primitives. Leveraging spectral cues for efficient initialization, \citet{zhang2025image} and \citet{meuleman2025fly} utilize frequency-based edge detection to guide the initial placement of primitives. Similarly, \citet{mallick2024taming3dgs} integrate a frequency-based term directly into the densification logic to prioritize high-frequency details with limited memory. While this approach improves the decision of \textit{when} to split, it still follows a conservative, uniform split into two children that does not address \textit{how many} splits are needed to adequately resolve local structure. Conversely, methods like Mip-Splatting \cite{yu2024mipsplatting} focus on anti-aliasing by introducing low-pass filters that constrain the minimum size of Gaussian primitives based on their screen-space footprint. While effective at eliminating artifacts, this approach tends to suppress high-frequency details rather than reconstruct them.\looseness=-1

Our approach differs fundamentally by addressing not only \textit{when} to densify, but \textit{how} to densify based on local structure. While Mip-Splatting \textit{suppresses} high frequencies to match the representation, we \textit{actively densify} the geometry to \textit{resolve} those frequencies. By computing the dominant local frequency using structure tensor analysis \cite{forstner1987fast,di1986note} with Laplacian scale space~\cite{lindeberg1994scale} and comparing it against each Gaussian's projection, we derive an explicit \textit{frequency violation metric}. This enables more principled densification decisions, producing sharper reconstructions in textured regions without information loss.\looseness=-1
\section{Method}
\label{sec:method}

Our approach addresses the limitations of heuristic densification in 3DGS by integrating explicit frequency analysis into the densification process. 
After recalling preliminaries on 3DGS and multiscale image structure analysis~(\cref{subsec:prelim}), we introduce a structure-aware supervision signal derived from multi-scale frequency analysis~(\cref{subsec:structure_tensor}), a frequency violation metric~(\cref{subsec:sampling_metric}), and a structure-aware densification strategy that proactively subdivides under-resolved Gaussians~(\cref{subsec:analytic_split}). A multiview consistency mechanism (\cref{subsec:multiview}) ensures robust densification decisions.

\subsection{Preliminaries}
\label{subsec:prelim}

\subsubsection{3D Gaussian Splatting}

3DGS represents a scene using a set of anisotropic 3D Gaussians. Each Gaussian is parameterized by its position $\boldsymbol{\mu} \in \mathbb{R}^3$, covariance matrix $\Sigma \in \mathbb{R}^{3 \times 3}$, opacity $\alpha \in [0, 1]$, and view-dependent color. The covariance $\Sigma$ is decomposed into a scaling vector $\mathbf{s} = (s_x, s_y, s_z) \in \mathbb{R}^3$ and rotation quaternion $\mathbf{q} \in \mathbb{R}^4$ (converted to a rotation matrix $R \in \mathbb{R}^{3 \times 3}$), yielding $\Sigma = RSS^T R^T$ where $S = \text{diag}(\mathbf{s})$. This parameterization allows independent control over the size of each Gaussian along its three principal axes.

For rendering, each 3D Gaussian is projected to 2D screen space via the view transformation $W \in \mathbb{R}^{3 \times 3}$ and projection Jacobian $J \in \mathbb{R}^{2 \times 3}$. 
We denote the projected mean by $\boldsymbol{\mu}_{2D} \in \mathbb{R}^2$.
The three principal axes of the 3D Gaussian also project to 2D, yielding axis vectors
\begin{equation}
\label{eq:projected_axes}
    \mathbf{v}_k = JWR\,S_k, \quad k \in \{x, y, z\},
\end{equation}
where $S_k$ is the $k$-th column of $S$. Each $\mathbf{v}_k \in \mathbb{R}^2$ represents the projected extent of the Gaussian along its $k$-th principal axis. These vectors will be used in \cref{subsec:sampling_metric} to compare Gaussian resolution against local texture frequency.

The color of a pixel is computed by alpha blending the depth-ordered projected Gaussians that overlap the pixel.
The loss $\mathcal{L}$ used to optimize the representation from posed image observations is a weighted combination of $\ell_1$ and structural similarity~\cite{wang2004image} losses.
 
3DGS periodically densifies the representation by splitting Gaussians with high view-space positional gradients $\nabla_{\boldsymbol{\mu}_{2D}} \mathcal{L}$. However, this heuristic is reactive and lacks awareness of the local signal frequency.

\subsubsection{Structure Tensor \& Local Frequency Analysis}
\label{subsec:freq_background}

The principle of our method is to ensure that the frequency content induced by the Gaussian primitives matches the local frequency content of the scene. 

To reason about local image structure, we make use of the structure tensor~\cite{di1986note,forstner1987fast}.
Given an image $I$ pre-smoothed with a Gaussian kernel $G_\sigma$, the structure tensor at scale $\sigma$ is a $2 \times 2$ per-pixel symmetric matrix that summarizes the local gradient distribution:
\begin{equation}
\label{eq:structure_tensor}
    S_\sigma = G_\rho * (\nabla I_\sigma \nabla I_\sigma^T)
\end{equation}
where $I_\sigma = G_\sigma * I$ is the pre-smoothed image and $\rho$ is the integration scale. In component form, this yields
$$
S_{xx} = \sum_c I_{x,c}^2, \quad S_{xy} = \sum_c I_{x,c} I_{y,c}, \quad S_{yy} = \sum_c I_{y,c}^2,
$$
where $I_{x,c} = \nabla_x(G_\sigma * I_c)$ and $I_{y,c} = \nabla_y(G_\sigma * I_c)$ are the smoothed gradients of color channel $c$, following the multi-channel formulation of \citet{di1986note}. $S_{xx}$ measures gradient energy in the horizontal direction, $S_{yy}$ in the vertical direction, and $S_{xy}$ captures the cross-correlation between directions, encoding orientation. The eigenvalues $\lambda_1, \lambda_2$ of $S_\sigma$ represent gradient energy along principal directions of local orientation: high eigenvalues indicate regions with significant high-frequency content such as edges.


\changed{
At a fixed analysis scale $\sigma$, the energy of the structure tensor reflects both image contrast and spatial frequency content. The dependence on contrast is immediate, since $S_\sigma$ is constructed from image gradients. To illustrate the role of frequency, consider a sinusoidal signal $\sin(2\pi \omega x)$: its derivative is $2\pi \omega \cos(2\pi \omega x)$, so the gradient magnitude scales as $|I_x| \propto \omega$, and the corresponding tensor entry satisfies $S_{xx} \propto \omega^2$. In practice, pre-smoothing with $G_\sigma$ acts as a low-pass filter that attenuates frequencies above $\omega \sim 1/(2\pi\sigma)$. Consequently, the trace $\operatorname{tr}(S) = S_{xx} + S_{yy}$ captures the band-limited gradient energy, aggregating contributions from a range of spatial frequencies with higher frequencies contributing more strongly (up to the cutoff induced by $\sigma$). This relationship underlies our multi-scale analysis (\cref{subsec:structure_tensor}).
}

\subsection{Multi-scale Structure Tensor Analysis}
\label{subsec:structure_tensor}

To estimate local frequency, we leverage the structure tensor (Section~\ref{subsec:freq_background}) in combination with Laplacian scale-space analysis~\cite{lindeberg1994scale,scheunders2002wavelet}. The structure tensor captures local gradient orientation and magnitude, but by itself does not indicate the spatial scale at which these gradients are most prominent. Since real scenes contain structures at multiple frequencies -- ranging from fine texture to coarse edges -- a single-scale tensor cannot distinguish between them.
We therefore introduce scale explicitly by analyzing image structure across scales.

\changed{
Specifically, given an RGB image $I$, we construct a Gaussian scale space using Gaussian blur kernels $G_{\sigma_l}$ with increasing standard deviations $\sigma_l = 1.5^l$, for $l \in {0, \ldots, L}$ (we use $L = 4$ in practice). We maintain full spatial resolution at all levels, yielding a set of images $\{I_l\}$.
Unlike image pyramids that perform additional downsampling~\cite{lowe2004sift}, this formulation preserves pixel correspondences across scales. At each level $l$, we compute per-pixel structure tensors $S_l$ using an integration scale $\rho_l = 3\sigma_l$.
The structure tensor $S_l$ encodes the combined effects of image contrast, band-limited frequency content, as well as orientation and anisotropy (Section~\ref{subsec:freq_background}), which we aim to disentangle.
}

\changed{
First, to remove the influence of overall signal magnitude, we normalize the tensor:
\begin{equation}
\hat{S}_l = \frac{S_l}{\operatorname{tr}(S_l) + \epsilon},
\end{equation}
where $\epsilon > 0$ ensures numerical stability. This normalization suppresses contrast- and frequency-dependent scaling captured by $\operatorname{tr}(S_l)$, and emphasizes orientation and anisotropy.
}

\changed{
Next, we estimate per-scale texture energy using a Laplacian scale-space formulation. The difference between adjacent blur levels acts as a band-pass filter that isolates image content within a specific frequency band. We define the corresponding energy as
\begin{equation}
E_l = \| I_{l-1} - I_l \|_2,
\end{equation}
computed per pixel. High values of $E_l$ indicate that significant texture energy is present at the corresponding scale.
}

\changed{
Finally, we aggregate the structure tensors, weighted by their corresponding energy and modulated by the squared frequency:
\begin{equation}
\bar{S} =
\frac{\sum_{l=0}^{L} E_l^{\gamma} \, \omega_l^2 \, \hat{S}_l }
{\sum_{l=0}^{L} E_l^{\gamma} + \epsilon},
\end{equation}
where $\gamma$ controls the sharpness of the weighting, emphasizing scales with higher energy; it was chosen empirically and set to $3.0$.
This aggregation emphasizes scales at which the local structure exhibits strong responses, reflecting both the dominant frequency content and the corresponding orientation and anisotropy.
The resulting tensor map $\bar{S}$ provides scale-adaptive guidance for structure-aware densification.
Fig.~\ref{fig:teaser} (left panel) visualizes $\bar{S}^{-1}$ as ellipses.
}

\subsection{Frequency Violation Metric}
\label{subsec:sampling_metric}

\changed{
We propose a per-Gaussian frequency violation metric $\eta$ that compares the size of projected scene Gaussians to the local texture wavelength observed in the input images, as estimated by the multiscale structure tensor $\bar{S}$ introduced above, thereby providing a more principled densification criterion than previous heuristics.
}

\changed{
To associate each scene Gaussian with a structure tensor from a training image, we uniformly sample $\bar{S}$ within the Gaussian's image-space footprint, average the samples to obtain a representative tensor, and then compute its principal eigenvalue $\lambda_1$.
The minimum local wavelength is given by $\Lambda_{\min} = 1/(\sqrt{\lambda_1} + \epsilon)$, representing the smallest spatial scale of detail that can be reliably captured within the Gaussian’s region. Higher eigenvalues correspond to more rapid spatial variation and thus finer texture.
}

\changed{
Our frequency violation metric is defined as the ratio of the magnitudes of the Gaussian's projected principal axis vectors $\mathbf{v}_k$ (\cref{eq:projected_axes}) to $\Lambda_{\min}$:}

\begin{equation}
\label{eq:our-eta}
\boxed{
    \eta_k = \frac{\|\mathbf{v}_k\|_2}{\Lambda_{\min}}.
}
\end{equation}
When $\eta > 1$, the Gaussian's projected extent exceeds the local texture wavelength, indicating potential under-resolution. Geometrically, this implies that a single Gaussian cannot accurately reconstruct higher-frequency spatial variations. This frequency violation directly correlates with reconstruction error and guides our adaptive densification strategy. The three-dimensional violation vector $\boldsymbol{\eta} = (\eta_x, \eta_y, \eta_z)$ enables anisotropic densification.


An alternative formulation projects the aggregated structure tensor onto each axis direction:
\begin{equation}
\label{eq:implementation_variants}
    \eta_k^{\text{(proj)}} = \sqrt{S_{xx} u_k^2 + 2 S_{xy} u_k v_k + S_{yy} v_k^2},
\end{equation}
where $(u_k, v_k)$ are the components of $\mathbf{v}_k$ \changed{from \cref{eq:projected_axes}}. This approach provides a stricter anisotropic criterion where each axis is evaluated against its directional frequency content. 
\changed{However, as shown in \cref{subsec:ablation}, this alternative yields lower-quality view synthesis results.}

\subsection{Multiview Consistency}
\label{subsec:multiview}

Single-view $\eta$ measurements may be noisy due to occlusion or view-dependent effects.
To avoid densifying Gaussians that are occluded or located far from visible surfaces, we aggregate observations across training views, \changed{maintaining per-axis counts $N$ for high ($\eta > 1$) and low ($\eta < 0.1$) responses, along with the maximum observed response $\eta_{\max}$.}

\changed{Densification is triggered only when a significant fraction of views report consistent violations, i.e., $\frac{N_{\text{high}}}{N_{\text{total}}} > \tau_{\text{split}}$.}

This ensures decisions are robust to transient errors.
Similarly, we prune Gaussians that consistently report low $\eta$ across views (indicating they cover smooth regions where smaller primitives are not needed) and have low opacity, \changed{i.e., $\frac{N_{\text{low}}}{N_{\text{total}}} > \tau_{\text{prune}}$ and $\alpha < \tau_{\alpha}$.}

\changed{We set the thresholds $\tau_{\text{split}} = \tau_{\text{prune}} = 0.8$ and $\tau_{\alpha} = 0.1$ in all our experiments.}\looseness=-1

\subsection{Structure-aware Densification}
\label{subsec:analytic_split}

Most previous methods~\cite{kerbl20233dgs,mallick2024taming3dgs,ren2026fastgs} split each Gaussian into two children with uniform shrinkage, requiring multiple iterations to achieve fine resolution. In addition, uniform shrinkage struggles with thin structures, as it fails to adapt the shape of anisotropic Gaussians. Our structure-aware densification instead determines the split factor directly from $\eta_{\max}$ (we use $\eta$ throughout this subsection to avoid notational clutter).

Recall that $\eta_k$ quantifies the factor by which the Gaussian’s extent exceeds the local wavelength along the $k$-th principal direction.
Ideally, the Gaussian should be split into approximately $\eta_k$ children along axis~$k$ so that each child matches the local frequency content.
\changed{However, directly setting the split factor to $n = \lceil \eta \rceil$ can lead to excessive splitting when $\eta$ is large or noisy. To improve robustness and to maintain memory efficiency, we instead employ a concave mapping that compresses $\eta$ in the high-value regime.
As analyzed in \cref{subsec:ablation}, we found empirically that the concave function
\begin{equation}
\label{eq:split-factor}
n = \lceil \sqrt{\eta} \rceil
\end{equation}
performs best among alternatives.}
We compute $n$ independently for each dimension and generate $n_x \times n_y \times n_z$ Gaussian children arranged on a regular grid with positions
\begin{equation}
    \boldsymbol{\mu}_{\text{child}}^{(i,j,k)} = \boldsymbol{\mu}_{\text{parent}} + R_{\text{parent}} \cdot (\mathbf{s}_{\text{parent}} \odot \mathbf{g}^{(i,j,k)}),
\end{equation}
where $R_{\text{parent}}$ is the parent primitive’s rotation matrix, $\mathbf{s}_{\text{parent}}$ denotes the parent's scaling vector, $\mathbf{g}^{(i,j,k)}$ represents the coordinates corresponding to indices $i, j, k$ on a regular grid in the unit cube with resolution $n_x \times n_y \times n_z$, and $\odot$ denotes element-wise multiplication.
The children's scaling vectors are defined as $\mathbf{s}_{\text{child}} = \mathbf{s}_{\text{parent}} \oslash (n_x, n_y, n_z)$, where $\oslash$ denotes element-wise division. 
Our approach allows reaching the necessary resolution in fewer densification cycles while avoiding excessive memory growth, thus handing over the job to gradient-based optimizers at an earlier stage for faster convergence.
\changed{Figure~\ref{fig:split-comp} illustrates the difference between conventional densification and our approach.}

\subsection{Training}
\label{subsec:training}

Our method integrates into the standard 3DGS training loop with minimal overhead. Beyond the standard structure-from-motion initialization, we additionally place Gaussians on the faces of the scene's bounding box to ensure complete geometric coverage. The multi-scale structure tensors $\bar{S}$ are precomputed for each training image using vectorized GPU operations, incurring \changed{a one-time overhead of 0.7 seconds and an additional memory cost equal to the input images’ footprint}.

During training, we perform online accumulation of frequency statistics: in each iteration, for visible Gaussians, we reuse the 2D covariance and intermediate buffers from the forward rendering pass to sample one point within each Gaussian’s footprint and compute $\eta$. This avoids additional projection or sampling overhead. We use an additional $\ell_2$ loss for faster convergence.
Our structure-aware densification occurs every 500 iterations. 
\changed{In addition, we apply the densification strategy of AbsGS~\cite{ye2024absgs} every 100 iterations, primarily to populate empty regions with primitives so that our structure-aware densification can fully leverage the existing primitives.}\looseness=-1

\section{Experiments}
\label{sec:results}

\subsection{Experimental Setup}
\label{subsec:setup}

We evaluate our method on three widely used benchmark datasets.
Mip-NeRF 360~\cite{barron2022mipnerf360} features unbounded indoor and outdoor scenes with complex central objects and multi-scale backgrounds.
Deep Blending~\cite{hedman2018deepblending} contains various indoor environments captured for image-based rendering.
Tanks \& Temples~\cite{knapitsch2017tanks} includes large-scale outdoor scenes with challenging geometric complexity and a substantially larger number of training views.
We report PSNR, SSIM~\cite{wang2004image}, and LPIPS~\cite{zhang2018unreasonable} metrics on novel views, along with the number of Gaussians ($N_{GS}$) to assess model compactness and total training time to measure optimization efficiency.

Following common practice in fast 3DGS methods such as FastGS \cite{ren2026fastgs} and 3DGS-LM~\cite{hollein20253dgs} that employ different configurations for various scene types, we apply dataset-dependent training settings tailored to their  characteristics.
We employ batch training with scene-dependent batch sizes: a batch size of 2 for indoor scenes from Mip-NeRF 360 and Deep Blending, and a batch size of 1 for outdoor scenes. The larger batch size for indoor scenes improves coverage of background regions across viewpoints within our limited iteration budget; while a batch size of 1 typically reconstructs foreground geometry well, it leads to noticeably lower quality in background regions. Although increasing the batch size to 2 doubles the per-iteration training time, we still observe a significant speed-up in convergence compared to baseline methods.
We further use different training iterations depending on the dataset characteristics: 3k iterations for Mip-NeRF360 and Deep Blending, and 7k iterations for Tanks \& Temples.
The extended training for Tanks \& Temples is necessary because this dataset contains significantly more training cameras and a wider camera distribution range, requiring additional optimization steps to achieve convergence.
Despite these variations, our method remains substantially faster than all baselines. 
All timing results and quantitative numbers reported in this section, including those for our method and all baselines, were measured on a single Nvidia H100 GPU.

\begin{table*}[t]
\caption{Quantitative comparisons on the Mip-NeRF360~\cite{barron2022mipnerf360}, Deep Blending~\cite{hedman2018deepblending}, and Tanks \& Temples~\cite{knapitsch2017tanks} datasets. We report optimization time (in seconds), PSNR, SSIM, LPIPS, the number of Gaussians ($N_{GS}$), and the number of training iterations. Our method achieves competitive or superior quality while requiring significantly fewer training iterations and substantially less training time. {Note that for Mip-NeRF360 indoor scenes and all Deep Blending scenes, we use a batch size of 2 to improve background convergence, which nearly doubles the per-iteration training time; nevertheless, our method remains substantially faster than all baselines. } The \colorbox{bestcolor}{best}, \colorbox{secondbestcolor}{second-best}, and \colorbox{thirdbestcolor}{third-best} performers are highlighted.}
\vspace{-0.3cm}
\label{tab:results}
\centering
\small
\renewcommand{\tabcolsep}{0.1cm}
\begin{tabular}{lrrrrrrrrrrrrrrrrrr}
\toprule
\multirow{2}{*}{Method} & \multicolumn{6}{c}{Mip-NeRF360} & \multicolumn{6}{c}{Deep Blending} & \multicolumn{6}{c}{Tanks \& Temples} \\
\cmidrule(lr){2-7} \cmidrule(lr){8-13} \cmidrule(lr){14-19}
 & Time$\downarrow$ & PSNR$\uparrow$ & SSIM$\uparrow$ & LPIPS$\downarrow$ & $N_{GS} \downarrow$ & It.$\downarrow$ & Time$\downarrow$ & PSNR$\uparrow$ & SSIM$\uparrow$ & LPIPS$\downarrow$ & $N_{GS} \downarrow$ & It.$\downarrow$ & Time$\downarrow$ & PSNR$\uparrow$ & SSIM$\uparrow$ & LPIPS$\downarrow$ & $N_{GS} \downarrow$ & It.$\downarrow$ \\
\midrule
3DGS  &
972.9 & 27.54 & 0.813 & 0.221 & 2.63M & 30k &
969.0 & 29.75 & \third{0.903} & \third{0.241} & 2.46M & 30k &
575.0 & 23.68 & \third{0.849} & \third{0.171} & 1.57M & 30k \\

Mini-Splat &
926.7 & 27.37 & \best{0.821} & 0.217 & \third{0.53M} & 30k &
704.0 & \second{29.97} & \second{0.907} & 0.243 & 0.56M & 30k &
495.5 & 23.41 & 0.843 & 0.182 & \third{0.30M} & 30k \\

Speedy-Splat  &
704.8 & 26.89 & 0.781 & 0.295 & \best{0.30M} & 30k &
581.0 & 29.46 & 0.899 & 0.271 & \second{0.25M} & 30k &
343.5 & 23.36 & 0.816 & 0.242 & \best{0.18M} & 30k \\

Taming-Bgt &
277.1 & 27.37 & 0.793 & 0.263 & 0.67M & 30k &
174.0 & 29.72 & 0.899 & 0.274 & \third{0.29M} & 30k &
172.5 & 23.90 & 0.836 & 0.209 & 0.32M & 30k \\

Taming-Big &
589.0 & \best{27.98} & \second{0.820} & \second{0.211} & 3.21M & 30k &
459.5 & 29.64 & 0.900 & \second{0.239} & 2.80M & 30k &
355.5 & \second{24.35} & \second{0.855} & \second{0.166} & 1.83M & 30k \\

DashG-Base &
323.9 & \third{27.70} & 0.814 & 0.217 & 2.16M & 30k &
201.0 & 29.68 & 0.900 & 0.250 & 1.99M & 30k &
243.5 & 23.95 & 0.840 & 0.178 & 1.33M & 30k \\

DashG-Big &
339.7 & 27.69 & \third{0.817} & 0.218 & 2.41M & 30k &
249.0 & 29.66 & 0.902 & 0.248 & 1.98M & 30k &
286.0 & 23.95 & 0.841 & 0.177 & 1.34M & 30k \\

FastGS-Base &
\second{143.7} & 27.55 & 0.797 & 0.261 & \second{0.40M} & 30k &
\second{116.5} & 29.81 & 0.900 & 0.270 & \best{0.22M} & 30k &
\second{126.0} & \third{24.12} & 0.838 & 0.211 & \second{0.24M} & 30k \\

FastGS-Big &
\third{208.9} & \second{27.96} & \second{0.820} & \third{0.216} & 1.16M & 30k &
\third{141.5} & \best{30.17} & \second{0.907} & 0.243 & 0.65M & 30k &
\third{156.0} & \best{24.41} & \second{0.855} & 0.175 & 0.54M & 30k \\

\textbf{Ours} &
\best{52.96} & 27.25 & \best{0.821} & \best{0.197} & 4.05M & 3k &
\best{41.22} & \third{29.89} & \best{0.910} & \best{0.238} & 1.81M & 3k &
\best{84.44} & 23.60 & \best{0.857} & \best{0.147} & 1.86M & 7k \\

\bottomrule
\end{tabular}
\end{table*}

\subsection{Comparison with Fast 3DGS Methods}
\label{subsec:comparison}

\paragraph{Baselines.}
We compare against the original 3DGS~\cite{kerbl20233dgs} as the  representative of reactive densification methods. 
To evaluate our method's efficiency against other acceleration strategies, we also compare against state-of-the-art fast 3DGS variants, including Mini-Splatting~\cite{fang2024minisplatting}, Speedy-Splat~\cite{hanson2025speedy}, Taming-3DGS~\cite{mallick2024taming3dgs}, DashGaussian~\cite{chen2025dashgaussian}, and FastGS~\cite{ren2026fastgs}.
For the fast variants, we include both their base and large/high-quality configurations where applicable to provide a comprehensive comparison.

\paragraph{Quantitative Results.}
As shown in Table~\ref{tab:results}, our method achieves the fastest training time while simultaneously obtaining the best SSIM and LPIPS scores across all three datasets, with competitive PSNR.
On Mip-NeRF360, we complete training in 53 seconds ($18\times$ faster than 3DGS, $2.7\times$ faster than FastGS-Base).
On Deep Blending, we finish in 41 seconds ($23\times$ faster than 3DGS, $2.8\times$ faster than FastGS-Base).
On Tanks \& Temples, we complete in 84 seconds ($6.8\times$ faster than 3DGS, $1.5\times$ faster than FastGS-Base).
Compared to FastGS-Big, which achieves the second-best LPIPS, our method improves LPIPS by 9\% on Mip-NeRF360 while being $4\times$ faster, and by 16\% on Tanks \& Temples with nearly $2\times$ speedup.
While FastGS-Base is the previously fastest baseline, we outperform its convergence speed by $2.7\times$ with 25\% better LPIPS on Mip-NeRF360, and by $1.5\times$ with 30\% better LPIPS on Tanks \& Temples.
These results demonstrate that our structure-aware densification not only accelerates training but also leads to superior perceptual quality.

\paragraph{Convergence Analysis.}
Figure~\ref{fig:curves_exp} illustrates the image quality metrics as a function of training time.
We include FastGS-Base-30k and FastGS-Big-30k as baselines using the official FastGS implementation. {FastGS is the state of the art on fast convergence of 3DGS; other baselines are omitted from this experiment due to the significant speed gap between their methods and FastGS (as demonstrated by FastGS).}
To validate that our speedup is not simply caused by position learning rate tuning (as shown in our ablation study), but actually stems from our structure-aware densification strategy, we additionally evaluate FastGS-Base-7k and FastGS-Big-7k variants. We run these variants for 7k iterations using their official configurations, with the only modification being the use of the same accelerated position learning rate schedule as ours (\texttt{position\_lr\_max\_steps} set to 3k for Mip-NeRF360 and Deep Blending, 7k for Tanks \& Temples).
Our method converges significantly faster than all baselines, faithfully capturing high-frequency details after only 1--2 densification splits.
On Mip-NeRF360 and Deep Blending, we reach reasonable quality as early as 30 seconds, which is reflected in the LPIPS curves.
This rapid convergence stems from our structure-aware densification strategy, which enables the representation to reach the necessary resolution early in the optimization process, bypassing the lengthy gradual densification stages characteristic of reactive gradient-based methods. While the FastGS-Base-7k variant shows faster early convergence than its 30k counterpart, it still suffers from the limitations of heuristic densification, as reflected in its blurry texture reconstructions in Figure~\ref{fig:curves_exp}. Notably, even with 30k iterations and significantly longer training time, FastGS-Big still fails to reach our LPIPS performance, demonstrating that our strategy captures high-frequency details both faster and more effectively.

\begin{figure*}[t]
    \centering
    \includegraphics[width=\linewidth]{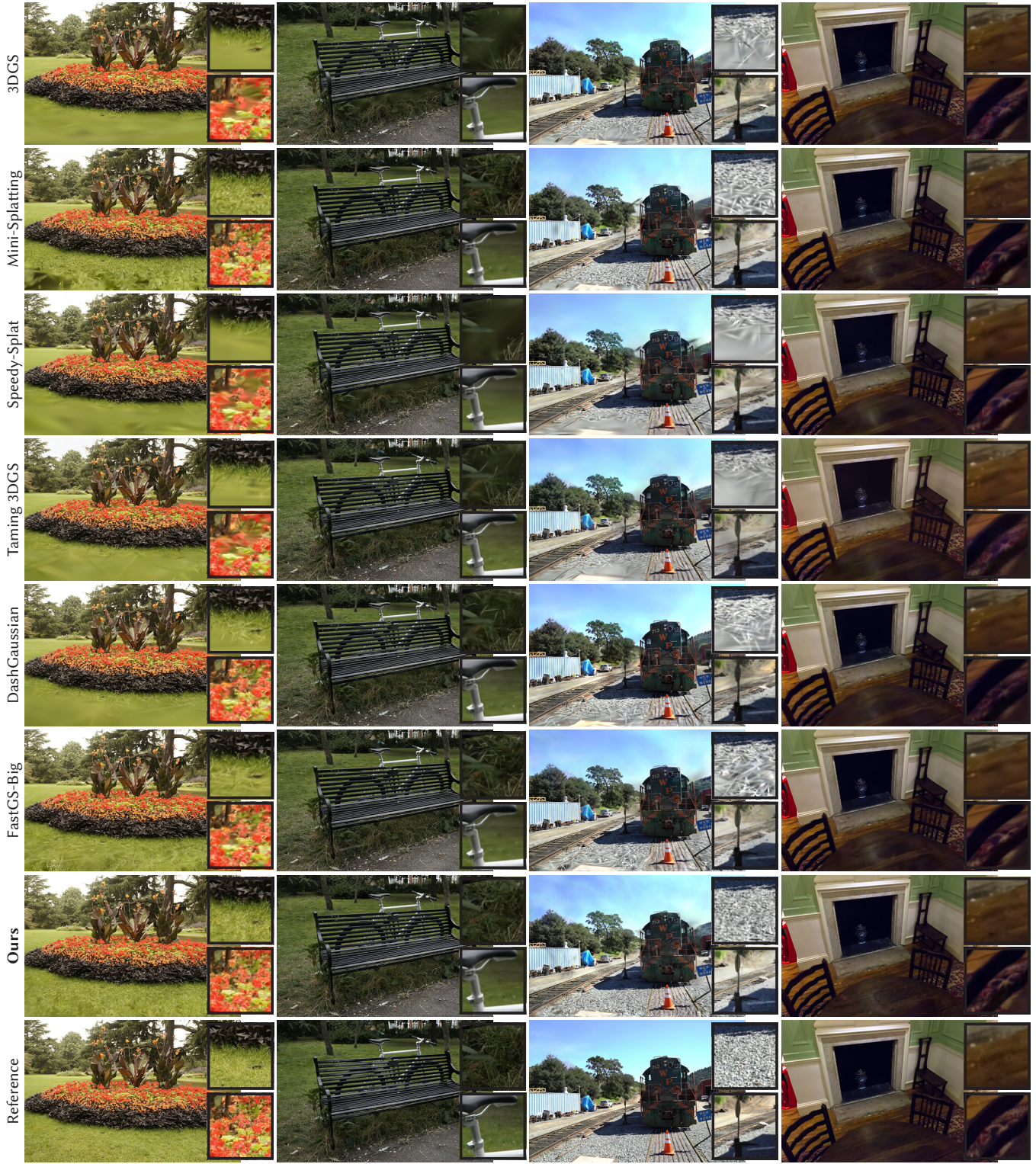}
    \caption{Qualitative results. We compare our approach with state-of-the-art fast 3DGS methods (rows) across multiple scenes (columns). Our method achieves a 3–20$\times$ speedup while simultaneously reconstructing significantly richer high-frequency details.}
    \label{fig:qualitative_results}
\end{figure*}

\paragraph{Qualitative Comparison.}
Figure~\ref{fig:qualitative_results} presents visual comparisons across multiple scenes.
Our method reconstructs high-frequency regions substantially better than competing approaches, particularly for fine-grained structures such as grass, leaves, and intricate textures.
This improvement is directly attributable to our structure-aware densification strategy, which explicitly aligns Gaussian primitives with local image structures based on frequency analysis.
While other methods struggle to resolve these detailed regions even after 30k iterations, our approach captures them accurately within 1.5 minutes.
\changed{For more qualitative results and an analysis of rendering speed and peak training VRAM, please refer to the supplemental.}

\subsection{Ablation Study}
\label{subsec:ablation}

\begin{table*}[t]
\caption{Ablations. We compare our full method against three variants: (1) $\eta^{\text{(proj)}}$, using a projection-based frequency violation metric; (2) \emph{30k LR Schedule}, using the original 30k position learning rate schedule instead of our accelerated schedule; (3) \emph{w/o MV Consistency}, disabling the multiview consistency mechanism.}
\vspace{-0.3cm}
\label{tab:ablation}
\centering
\small
\renewcommand{\tabcolsep}{0.1cm}
\begin{tabular}{lrrrrrrrrrrrrrrr}
\toprule
\multirow{2}{*}{Method} & \multicolumn{5}{c}{Mip-NeRF360} & \multicolumn{5}{c}{Deep Blending} & \multicolumn{5}{c}{Tanks \& Temples} \\
\cmidrule(lr){2-6} \cmidrule(lr){7-11} \cmidrule(lr){12-16}
 & Time$\downarrow$ & PSNR$\uparrow$ & SSIM$\uparrow$ & LPIPS$\downarrow$ & $N_{GS} \downarrow$ & Time$\downarrow$ & PSNR$\uparrow$ & SSIM$\uparrow$ & LPIPS$\downarrow$ & $N_{GS} \downarrow$ & Time$\downarrow$ & PSNR$\uparrow$ & SSIM$\uparrow$ & LPIPS$\downarrow$ & $N_{GS} \downarrow$ \\
\midrule
$\eta^{\text{(proj)}}$ &
\best{46.97} & \third{27.20} & \third{0.819} & \third{0.203} & \best{3.18M} &
\best{38.17} & \best{29.89} & \best{0.910} & \third{0.242} & \best{1.35M} &
\best{70.59} & \second{23.56} & \third{0.855} & \third{0.159} & \best{1.86M} \\

30k LR Schedule &
\second{52.75} & 26.01 & 0.766 & 0.261 & \second{3.92M} &
\second{40.38} & 28.39 & 0.878 & 0.308 & \second{1.59M} &
\second{79.64} & 22.94 & 0.815 & 0.203 & \third{2.28M} \\

w/o MV Consistency &
56.48 & \best{27.30} & \best{0.822} & \best{0.196} & 4.56M &
41.77 & \third{29.80} & \best{0.910} & \best{0.237} & 2.30M &
89.94 & \third{23.46} & \best{0.857} & \best{0.146} & 2.84M \\

\textbf{Ours} &
\third{52.96} & \second{27.25} & \second{0.821} & \second{0.197} & 4.05M &
\third{41.22} & \best{29.89} & \best{0.910} & \second{0.238} & \third{1.81M} &
\third{84.44} & \best{23.60} & \best{0.857} & \second{0.147} & \best{1.86M} \\

\bottomrule
\end{tabular}
\end{table*}

We analyze the contribution of several components through controlled experiments, all listed in Table~\ref{tab:ablation}.

\paragraph{$\eta$ vs.\ $\eta^{\text{(proj)}}$}
\changed{The projection-based metric $\eta^{\text{(proj)}}$ (\cref{eq:implementation_variants})} applies a stricter three-channel anisotropic criterion where each axis is evaluated against its directional frequency content. In contrast, our metric $\eta$ (\cref{eq:our-eta}) compares all axes against the same scalar wavelength, making it more likely to trigger densification on any axis that exceeds this global limit. While $\eta^{\text{(proj)}}$ is theoretically more precise, we observe that $\eta$ achieves better LPIPS scores in practice. We hypothesize that during the computation of $\eta$, the Gaussian rotations are not yet well-optimized, leading to inaccurate axis-direction alignment. The scalar wavelength approach compensates for this by applying a more uniform densification criterion that does not depend on correct axis orientation, giving the gradient-based optimization more capacity to subsequently fine-tune both rotation and position.\looseness=-1

\paragraph{Position Learning Rate Schedule}
Tuning the maximum steps for the position learning rate scheduler to match our accelerated training regime is essential. 
Using the original 30k schedule with our reduced training budget results in significant quality degradation across all metrics, as the position learning rate remains too high throughout most of training, preventing the Gaussians from settling into stable configurations.

\paragraph{Multiview Consistency}
Disabling the multiview consistency mechanism leads to marginally better reconstruction quality in some cases, as the method can respond more aggressively to single-view frequency violations. However, this comes at the cost of a 12--53\% increase in Gaussian count across datasets,

with correspondingly longer training times. We therefore use multiview consistency by default, as it provides a favorable trade-off between quality and model compactness.

\changed{
\paragraph{Split Factor.}

Our split factor, defined in \cref{eq:split-factor}, employs a concave mapping to compress the raw frequency violation $\eta$ and prevent excessive splitting due to outliers.
To evaluate this design choice, we consider a family of mappings of the form $n = \lceil \eta^{p} \rceil$.
Table~\ref{tab:split_factor} reports results for different choices of $p$ across all scenes in the Mip-NeRF360 dataset.
Training time and Gaussian count increase with $p$, while image quality largely saturates at $p = 0.5$ (i.e., the square root), which we adopt as our default.

\begin{table}[h]
\caption{\changed{Effect of concave mappings of the split factor on quality.}}
\vspace{-0.2cm}
\label{tab:split_factor}
\centering
\renewcommand{\tabcolsep}{0.12cm}
\begin{tabular}{lrrrrr}
\toprule
$p$ & Time$\downarrow$ & PSNR$\uparrow$ & SSIM$\uparrow$ & LPIPS$\downarrow$ & $N_{GS}\downarrow$ \\
\midrule
0.25              & 45.3 & 27.12 & 0.817 & 0.209 & 3.1M \\
0.5 (Ours)        & 53.0 & 27.25 & 0.821 & 0.197 & 4.1M \\
0.75              & 72.4 & 27.15 & 0.819 & 0.194 & 5.7M \\
\bottomrule
\end{tabular}
\end{table}
}

\changed{Additional ablations---including a comparison under matched Gaussian count, the effect of extended training, and the effect of the AbsGS densification component---are provided in the supplementary material.}

\subsection{Limitations}
\label{subsec:limitations}

\paragraph{PSNR Performance}
While our method achieves the best SSIM and LPIPS scores across all three benchmark datasets, our PSNR is slightly lower than some competing methods. This is primarily due to our significantly reduced training iterations (3k vs. 30k in standard 3DGS). With fewer optimization steps, the Gaussian positions do not fully converge to their optimal locations, resulting in small positional offsets that manifest as high-frequency rendering errors. These sub-pixel misalignments penalize PSNR more heavily than perceptual metrics, which are more robust to minor spatial shifts. Our strong performance on SSIM and LPIPS indicates that the reconstructions are perceptually superior despite the lower PSNR.

\paragraph{Gaussian Count}
Our frequency-aware densification strategy tends to produce a higher number of Gaussians compared to budget-constrained methods. This over-densification arises because our metric $\eta$ is designed to match the \textit{highest} local frequency within each Gaussian's footprint, ensuring that every child primitive can faithfully represent this peak frequency content. However, not all regions within a Gaussian require such fine resolution---only the edges or texture boundaries do. In practice, our stochastic jitter sampling and the conservative split factor $n$ 
partially compensate for this tendency, reducing unnecessary splits.

\changed{A further limitation is that $\eta$ evaluates each Gaussian in isolation, ignoring the collective contribution of overlapping primitives; as a result, densification may be triggered in regions where existing primitives already provide sufficient frequency support, contributing to over-densification.}

Ideally, one would implement a progressive and edge-count-based splitting strategy that subdivides Gaussians iteratively only along detected discontinuities, yielding a more compact representation. We explored this approach but found that it leads to longer convergence times. The underlying trade-off is that gradual densification introduces new Gaussians later in training, leaving insufficient iterations for these primitives to be fully optimized. Consequently, training time is effectively wasted on oversized Gaussians if they are not split early enough. We leave the development of a more compact densification strategy that still enables early splitting---without sacrificing training efficiency---as promising future work.

\paragraph{Stochastic Jitter Sampling}
Our frequency violation estimation relies on stochastic jitter sampling within each Gaussian's footprint to robustly estimate local frequency requirements. While this approach improves coverage and reduces sensitivity to edge alignment, it introduces minor variance across training runs. In practice, this variance is negligible and does not significantly affect final reconstruction quality.

\section{Conclusion}
\label{sec:conclusion}

We have presented a structure-aware 3D Gaussian Splatting framework that addresses the densification bottleneck from a signal-processing perspective. Unlike previous heuristics that only identify \textit{where} to split---requiring multiple successive cycles before reaching the necessary resolution---our frequency violation metric $\eta$ reveals not just the location but also \textit{how} to split (anisotropically, matching local feature orientation) and \textit{how many} splits are required per axis. By leveraging multi-scale frequency analysis combining structure tensors with Laplacian scale space, our method enables early analytic densification that bypasses the lengthy iterative ``split--train--split'' cycles. Combined with a multiview consistency mechanism, our method achieves state-of-the-art perceptual quality across three benchmark datasets while reducing training time by up to 23$\times$ compared to the original 3DGS approach.

Several promising directions remain for future exploration. Replacing the Adam~\cite{kingma2014adam} optimizer with second-order optimization methods 
could further accelerate convergence by providing more accurate curvature information for position and shape parameters. Integrating adaptive budget control strategies that dynamically balance primitive count against reconstruction quality would enable deployment on memory-constrained devices while preserving our fast convergence properties. Extending our frequency-aware framework to handle dynamic scenes and temporal consistency also presents an exciting avenue for future research.

\bibliographystyle{ACM-Reference-Format}
\bibliography{ref}

\begin{figure*}[t]
\centering
\begin{minipage}[t]{0.49\textwidth}
    \vspace{0pt}
    \centering
    \includegraphics[width=\linewidth]{fig/curves_03.ai}
    \vspace{-0.6cm}
    \caption{Image quality versus training time, averaged over two scenes from Mip-NeRF360 (\emph{a}), Deep Blending (\emph{b}), and Tanks \& Temples (\emph{c}). Our approach reaches high quality significantly faster than all baselines. The flags denote results corresponding to the final state of a method, whereas crossed-out flags indicate that additional iterations are required beyond those shown.}
    \label{fig:curves_exp}
\end{minipage}\hfill
\begin{minipage}[t]{0.47\textwidth}
    \vspace{0pt}
    \centering
    \includegraphics[width=\linewidth]{fig/split_comparison_01.ai}
    \vspace{-0.6cm}
    \caption{
        \changed{Comparison of conventional densification and our approach.
        The top row shows a training image (left) and a rendering of a 3DGS model at an early training stage (right).
        The bottom row visualizes the effect of a single densification step applied to this model. To highlight individual Gaussian primitives, we overlay the renderings with random colors per Gaussian.
        Conventional densification (bottom left) fails to resolve high-frequency structures, requiring multiple densification stages interleaved with gradient-based optimization to achieve high fidelity. 
        In contrast, our method (bottom right) leverages frequency information from the input image to guide anisotropic splitting, enabling the representation to capture fine structural details early in training.}
    }
    \label{fig:split-comp}
\end{minipage}
\end{figure*}

\end{document}


\title{Supplementary Material: Faster 3D Gaussian Splatting Convergence via Structure-Aware Densification}

\maketitle

\section*{Properties of the Structure Analysis}
\label{subsec:supp_structure_properties}

Our multiscale structure-tensor analysis is built on multiscale gradients, which in their raw form encode both local contrast and frequency.
The normalizations applied to the per-scale tensors (see Eq.~5 in the main paper) effectively remove amplitude effects, largely isolating the frequency component.
To quantify robustness, we measure the average relative change in the aggregated tensor map $\bar{S}_{\mathrm{GT}}$ under several image perturbations.

\paragraph{Contrast variations.}
Modifying the image exposure by a factor of $0.5$ or $1.5$ results in an average change of only $5\%$ or $8\%$ in $\bar{S}_{\mathrm{GT}}$, confirming that the analysis is robust to global contrast shifts.

\paragraph{Noise.}
Due to the image pre-smoothing step (see L319 of the main paper), additive image noise likewise has little effect: adding Gaussian noise with unit standard deviation results in an average change of only $7\%$.

\paragraph{Other image degradations.}
As with any reconstruction algorithm, our approach is more sensitive to degradations that alter high-frequency content.
Sharpening by $1.5\times$/$3\times$ leads to changes of $12\%$/$40\%$, and JPEG compression at quality level~80 results in a change of $12\%$.

\paragraph{Semantic sensitivity.}
Our structure analysis responds strongly to material reflectance, specular highlights, and shadow boundaries.
This behavior is desirable: these scene attributes are integral components of what a radiance field should faithfully represent, and their presence in $\bar{S}_{\mathrm{GT}}$ appropriately triggers densification in visually demanding regions.

\section*{Rendering Speed and Peak VRAM}
\label{subsec:supp_rendering}

Beyond training efficiency, we evaluate the inference-time performance of the resulting models.
Table~\ref{tab:rendering} reports rendering speed (frames per second, measured on an H100 GPU) and peak GPU memory consumption during training for a representative subset of methods, averaged across all datasets.
Our method achieves rendering rates comparable to 3DGS~\cite{kerbl20233dgs} (251 vs.\ 296\,FPS), confirming that the larger Gaussian count does not translate into a proportional rendering overhead.
The higher peak VRAM relative to 3DGS~\cite{kerbl20233dgs} and FastGS-Big~\cite{ren2026fastgs} reflects the additional memory required for our structure analysis and multiview consistency buffers during training, not at inference time.

\begin{table}[h]
\caption{Rendering speed and peak training VRAM on an H100 GPU.}
\vspace{-0.2cm}
\label{tab:rendering}
\centering
\small
\renewcommand{\tabcolsep}{0.15cm}
\begin{tabular}{lrr}
\toprule
Method & FPS$\uparrow$ & VRAM (GB)$\downarrow$ \\
\midrule
3DGS         & 296 &  9.9 \\
FastGS-Big   & 415 &  6.6 \\
Ours         & 251 & 13.0 \\
\bottomrule
\end{tabular}
\end{table}

\section*{Comparison under Matched Gaussian Count}
\label{subsec:supp_matched_gaussians}

Table~\ref{tab:matched_gaussians} ensures a fair comparison with FastGS~\cite{ren2026fastgs} by matching the final Gaussian count (4.1M) via an increased FastGS split factor. We evaluate this matched-capacity baseline under both its original 30k training iterations and our accelerated 3k schedule. While FastGS requires 30k iterations—and thus 10x more training time—to nearly match our quality at this Gaussian count, its performance degrades substantially when restricted to our reduced 3k iteration budget and learning rate schedule. This proves that our method's superiority stems from our fundamental formulation changes, rather than merely benefiting from a higher Gaussian count.

\begin{table}[h]
\caption{Comparison under matched Gaussian count ($\approx$4.1M) on Mip-NeRF360.}
\vspace{-0.2cm}
\label{tab:matched_gaussians}
\centering
\small
\renewcommand{\tabcolsep}{0.12cm}
\begin{tabular}{lrrrr}
\toprule
Method & Time$\downarrow$ & PSNR$\uparrow$ & SSIM$\uparrow$ & LPIPS$\downarrow$ \\
\midrule
FastGS (30k)     & 502.2 & 27.19 & 0.809 & 0.194 \\
FastGS (3k)      &  30.3 & 24.30 & 0.750 & 0.278 \\
Ours (3k)        &  53.0 & 27.25 & 0.821 & 0.197 \\
\bottomrule
\end{tabular}
\end{table}

\section*{Extended Training}
\label{subsec:supp_extended_training}

Table~\ref{tab:extended_training} demonstrates the rapid convergence of our method by detailing the effects of training beyond our default 3k iterations on the Mip-NeRF360~\cite{barron2022mipnerf360} dataset. While extending the training duration yields moderate overall improvements in PSNR and LPIPS, our method effectively converges very early. Notably, scene-specific analysis reveals that for outdoor scenes, there is no PSNR gap even at just 3k iterations. For indoor scenes, the remaining performance gap becomes negligible when training is extended to 30k iterations.

\begin{table}[h]
\caption{Effect of training duration on Mip-NeRF360.}
\vspace{-0.2cm}
\label{tab:extended_training}
\centering
\small
\renewcommand{\tabcolsep}{0.12cm}
\begin{tabular}{lrrrrr}
\toprule
\ Iterations & Time$\downarrow$ & PSNR$\uparrow$ & SSIM$\uparrow$ & LPIPS$\downarrow$ & $N_{GS}\downarrow$ \\
\midrule
3k (ours) &  53.0 & 27.25 & 0.821 & 0.197 & 4.1M \\
15k       & 194.4 & 27.55 & 0.821 & 0.186 & 4.1M \\
30k       & 295.3 & 27.48 & 0.818 & 0.184 & 4.1M \\
\bottomrule
\end{tabular}
\end{table}

\section*{Progressive Ablation}
Table~\ref{tab:progressive_ablation} reports a step-by-step ablation on all scenes of the Mip-NeRF360 dataset.
Starting from a base 3DGS model trained for 3k iterations, we sequentially add our 3k LR schedule, our structure-aware densification, and our multiview consistency. Each component contributes meaningfully to the final result.

\begin{table}[h]
\caption{Progressive ablation.}
\vspace{-0.2cm}
\label{tab:progressive_ablation}
\centering
\small
\renewcommand{\tabcolsep}{0.12cm}
\begin{tabular}{lrrrrr}
\toprule
Method & Time$\downarrow$ & PSNR$\uparrow$ & SSIM$\uparrow$ & LPIPS$\downarrow$ & $N_{GS}\downarrow$ \\
\midrule
Base                        & 33.2 & 25.84 & 0.748 & 0.296 & 1.5M \\
+ 3k LR Schedule            & 33.6 & 26.77 & 0.793 & 0.256 & 1.4M \\
+ Our Densification         & 54.4 & 27.31 & 0.822 & 0.196 & 4.5M \\
+ MV Consistency     & 53.0 & 27.31 & 0.821 & 0.197 & 4.0M \\
\bottomrule
\end{tabular}
\end{table}

\section*{AbsGS Densification}
\label{subsec:supp_absgs_warmup}

As described in the method, we apply the densification strategy of AbsGS~\cite{ye2024absgs} every 100 iterations to populate empty regions with primitives, ensuring that our structure-aware densification can fully leverage the existing primitive coverage.
Table~\ref{tab:absgs_warmup} shows that this component is not fundamental to our approach: removing it yields a slight reduction in Gaussian count and a small quality decrease (LPIPS diff of 0.005), but the method remains competitive.

\begin{table}[h]
\caption{Effect of the AbsGS densification component on Mip-NeRF360.}
\vspace{-0.2cm}
\label{tab:absgs_warmup}
\centering
\small
\renewcommand{\tabcolsep}{0.12cm}
\begin{tabular}{lrrrrr}
\toprule
Method & Time$\downarrow$ & PSNR$\uparrow$ & SSIM$\uparrow$ & LPIPS$\downarrow$ & $N_{GS}\downarrow$ \\
\midrule
w/ AbsGS densif.   & 52.96 & 27.25 & 0.821 & 0.197 & 4.05M \\
w/o AbsGS densif.  & 51.38 & 27.12 & 0.820 & 0.202 & 3.51M \\
\bottomrule
\end{tabular}
\end{table}

\section*{Effect of Initialization}
\label{subsec:supp_initialization}

Beyond the standard structure-from-motion point cloud, we additionally place Gaussians on the faces of the scene's bounding box to ensure complete geometric coverage.
To assess its contribution, we run an ablation in which this initialization is omitted while keeping all other components unchanged.
All quality metrics are near-identical without it (LPIPS difference of $0.002$).
Nevertheless, our initialization yields slight qualitative improvements in peripheral scene regions and consistently leads to a smaller total Gaussian count, as primitives are more evenly distributed from the outset.

\section*{Difference from DashGaussians}
\label{subsec:supp_dashgaussians}

DashGaussians~\cite{chen2025dashgaussian} and our method are complementary in motivation but operate on different timescales.
Although both approaches leverage multi-scale reasoning, our method establishes a scale-optimal Gaussian model early in training---resolving frequency violations in the first few densification steps---whereas DashGaussians progressively refines the representation over a longer schedule.
A naive combination of the two strategies would therefore not straightforwardly yield a benefit: inserting our early, proactive densification into DashGaussians' progressive pipeline disrupts its intended schedule, while appending DashGaussians' later-stage refinement after our method has already converged adds unnecessary training time without quality gain.
We therefore leave a more principled integration as future work.

\bibliographystyle{ACM-Reference-Format}
\bibliography{ref}